\documentclass[10pt,twocolumn,letterpaper]{article}
\usepackage[T1]{fontenc}
\usepackage{cvpr} 
\usepackage{array}
\usepackage{multirow}
\usepackage{placeins} 

\usepackage{cuted} 

\usepackage{currfile} 

\usepackage{caption} 

\usepackage{microtype}

\newcommand{\actionlabel}[1]{``\texttt{#1}''}
\newcommand{\actiondescription}[1]{``\texttt{#1}''}
\definecolor{cvprblue}{rgb}{0.21,0.49,0.74}
\usepackage[pagebackref,breaklinks,colorlinks,allcolors=cvprblue]{hyperref}
\hypersetup{pdfauthor={Siyu Zhu, Feng Ding, Zixiao Jin, Kaizhi Li, Xin Dong, Haisheng Fu, Yuchen Zou, Guohe Zhang, Jie Liang}}
\def\confName{CVPR}
\def\confYear{2027}
\title{Crafting the Difference, Grounding the Evidence: \\Discriminative Concept Learning for Zero-Shot Skeleton Action Recognition}
\author{%
\parbox{\dimexpr\textwidth-2\tabcolsep\relax}{\centering
\fontsize{10}{11}\selectfont
Siyu Zhu\textsuperscript{1,2}\quad
Feng Ding\textsuperscript{3}\quad
Zixiao Jin\textsuperscript{1,4}\quad
Kaizhi Li\textsuperscript{1,5}\quad
Xin Dong\textsuperscript{1,6}\\
Haisheng Fu\textsuperscript{7}\quad
Yuchen Zou\textsuperscript{2}\quad
Guohe Zhang\textsuperscript{2}\quad
Jie Liang\textsuperscript{1}\thanks{Corresponding author.}\\[2pt]
{\fontsize{8.5}{9.5}\selectfont
\textsuperscript{1}Eastern Institute of Technology, Ningbo\quad
\textsuperscript{2}Xi'an Jiaotong University\\
\textsuperscript{3}Simon Fraser University\quad
\textsuperscript{4}China University of Geoscience\quad
\textsuperscript{5}Shanghai Jiaotong University\\
\textsuperscript{6}University of Science and Technology of China\quad
\textsuperscript{7}University of British Columbia}
}}
\date{}

\makeatletter
\patchcmd{\@maketitle}{\vspace*{24pt}}{\vspace*{0pt}}{}{%
  \PackageError{craft}{Could not set title-to-author spacing}{}}
\patchcmd{\@maketitle}{\vskip .5em}{\vskip 0pt}{}{%
  \PackageError{craft}{Could not set spacing below the author block}{}}
\patchcmd{\@maketitle}{\vspace*{12pt}}{\vspace*{0pt}}{}{%
  \PackageError{craft}{Could not set spacing below the title block}{}}
\makeatother

\makeatletter
\AtBeginDocument{%
  \g@addto@macro\normalsize{%
    \addtolength{\abovedisplayskip}{-2pt}%
    \addtolength{\belowdisplayskip}{-2pt}}%
  \normalsize
}
\makeatother
\begin{document}
\raggedbottom 
\maketitle

\begin{abstract}
LLM-generated descriptions enrich zero-shot skeleton-based action recognition but repeat cues shared by confusable actions. Their contextual semantics also lack explicit links to skeleton features. To address these challenges, we propose \textbf{CRAFT}, which couples \textbf{C}oncept \textbf{R}efinement \textbf{a}nd \textbf{F}eature-grounded \textbf{T}okens. Hard-negative-conditioned generation and contrastive scoring identify concepts that distinguish each action from its closest rivals. Unique-QDPP selects complementary descriptions within each semantic slot and prevents their reuse across slots. This bank supervises learnable concept tokens that read motion and context cues from each skeleton. On fixed NTU-60 and NTU-120 splits, \textbf{CRAFT} improves five baselines spanning four recognition paradigms. ZSL accuracy and GZSL harmonic mean increase on every baseline-split pair, by 5.4 and 5.9 percentage points on average. Random-split experiments extend these gains to three datasets, including PKU-MMD. Code will be released after acceptance.
\end{abstract}

\section{Introduction}
\label{sec:introduction}

Human action recognition~\cite{sun2023modalities,zhu2026topology} supports healthcare monitoring~\cite{saggese2019learning}, security~\cite{sultani2018realworld}, and human-computer interaction~\cite{molchanov2016online}. Skeleton sequences compactly encode joint geometry and motion while limiting exposure of detailed appearance information~\cite{yan2018spatial,moon2023anonymization,gong2026federated}. Fully supervised methods learn from labeled sequences~\cite{shi2019twostream,cheng2020shift,chen2021channelwise,pang2022igformer}, but recognize only predefined classes.

\begin{figure}[t]
    \centering
    \includegraphics[width=0.8\columnwidth]{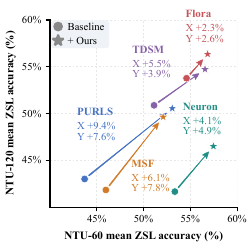}
    \caption{Mean fixed-split ZSL accuracy (\%) per dataset. Arrows show absolute gains from adding \textbf{CRAFT} to each baseline.}
    \label{fig:zsl-teaser}
\end{figure}

Zero-shot methods recognize unseen actions through skeleton-text alignment~\cite{jasani2019skeleton,gupta2021synse}, whereas one-shot methods require a labeled exemplar for each novel class~\cite{liu2026skelhcc}. LLM-generated descriptions supply cues beyond action labels~\cite{zhu2024purls,chen2024star}, and pretrained CLIP~\cite{radford2021clip} text encoders map them into semantic representations~\cite{chen2025neuron,wu2025fsvae,do2025tdsm}. Semantic refinement and contextual modeling strengthen this alignment~\cite{xu2025infocpl,chen2026flora,wang2026skeletoncontext}. Training-free cache adaptation further improves unseen-action recognition~\cite{zhu2025skeletoncache}.

Existing zero-shot methods generate descriptions from prompts centered on individual action labels~\cite{zhou2023smie,zhu2024purls,chen2024star,xu2025infocpl,chen2025neuron,wu2025fsvae,kuang2026dvta,wang2026skeletoncontext}. These prompts cover body-part motion, temporal stages, and action context, broadening semantic coverage within each class. Descriptions of confusable actions often repeat the same motion patterns, obscuring the cues that distinguish them. For example, \actiondescription{both hands move repeatedly in front of the torso} fits \actionlabel{clapping} and \actionlabel{rub two hands together} equally well. The descriptions \actiondescription{the hands repeatedly approach and separate} and \actiondescription{the hands remain close during short back-and-forth movements} capture their distinct relative hand motions.
These descriptions guide skeleton-text alignment, collectively or through random sampling~\cite{zhu2024purls,chen2024star,wang2026skeletoncontext}. Context introduces a further challenge: skeleton sequences encode joint geometry and motion, while descriptions may refer to objects and scenes~\cite{wang2026skeletoncontext}. For \actionlabel{writing}, \actiondescription{using a pen on paper} describes an interaction whose connection to skeleton motion remains implicit. Discriminative concepts should distinguish confusable actions while retaining complementary motion and context cues. These concepts should provide explicit learning targets for skeleton representations that capture those distinctions.

To tackle these challenges, we propose \textbf{CRAFT}, which couples \textbf{C}oncept \textbf{R}efinement \textbf{a}nd \textbf{F}eature-grounded \textbf{T}okens (Fig.~\ref{fig:main-framework}). Hard-negative-conditioned generation compares each action with its most confusable neighbors. Contrastive quality scoring favors candidates that separate the target from these neighbors. Unique-QDPP then selects complementary descriptions within each semantic slot and prevents duplicates across slots. The selected concepts supervise learnable tokens through concept-supervised skeleton prompt learning. These tokens read features from each input skeleton, connecting semantic distinctions to motion and context evidence.

We integrate compatible \textbf{CRAFT} framework into five baselines covering variational~\cite{li2023msf}, alignment-based~\cite{zhu2024purls,chen2025neuron}, diffusion-based~\cite{do2025tdsm}, and flow-based~\cite{chen2026flora} recognition. On fixed NTU-60 and NTU-120 splits, CRAFT improves all 40 baseline-split pairs in zero-shot learning (ZSL) and all 20 pairs in generalized ZSL (GZSL). The mean gains are 5.4 percentage points in ZSL accuracy (Fig.~\ref{fig:zsl-teaser}) and 5.9 percentage points in GZSL harmonic mean. Random-split experiments with TDSM and Flora show gains in every setting on NTU-60, NTU-120, and PKU-MMD.
Our contributions are as follows:
\begin{itemize}
    \item We propose discriminative concept construction that conditions generation on confusable actions and couples contrastive quality scoring with within-slot diversity and cross-slot uniqueness in Unique-QDPP.
    \item We propose concept-supervised skeleton prompt learning that turns selected concepts into slot-specific learning targets. These targets supervise learnable tokens to read input-dependent skeleton features, linking semantic distinctions to action evidence.
    \item As a plug-and-play enhancement, \textbf{CRAFT} improves five baselines across four recognition paradigms on fixed NTU splits. Random-split gains with TDSM and Flora extend across NTU-60, NTU-120, and PKU-MMD.
\end{itemize}

\section{Related Work}
\label{sec:related_work}

\subsection{Zero-Shot Skeleton-Based Action Recognition}

Early work connects skeleton features with class-label embeddings in a joint pose-language space~\cite{jasani2019skeleton}. SynSE learns syntactically guided generative embeddings~\cite{gupta2021synse}, while MSF uses multi-semantic fusion~\cite{li2023msf}. SA-DVAE disentangles features~\cite{li2024sadvae}, and FS-VAE combines frequency decomposition with multilevel semantic alignment~\cite{wu2025fsvae}. SMIE estimates and maximizes mutual information~\cite{zhou2023smie}, while DVTA uses deep metric learning for skeleton-text alignment~\cite{kuang2026dvta}. TDSM uses text-conditioned denoising~\cite{do2025tdsm}, and Flora uses flow-based classification~\cite{chen2026flora}. These methods develop feature representations and recognition objectives. \textbf{CRAFT} contributes discriminative alignment targets by constructing concepts that guide what skeleton-side tokens learn to express.

\subsection{Text-Driven Semantic Modeling}

PURLS~\cite{zhu2024purls} and STAR~\cite{chen2024star} use body-part motion descriptions to broaden semantic coverage. Neuron~\cite{chen2025neuron} uses multi-turn spatial and temporal prompting, while SkeletonContext~\cite{wang2026skeletoncontext} incorporates object and environment context. InfoCPL~\cite{xu2025infocpl} selects relevant cues from diverse descriptions according to the input skeleton. Flora~\cite{chen2026flora} refines encoded descriptions by aggregating semantic features from neighboring action classes. These methods refine semantic coverage, input relevance, or encoded features. \textbf{CRAFT} builds class comparisons into generation and selection to form concepts that complement one another within classes and separate confusable actions.

\begin{figure*}[!t]
    \centering
    \includegraphics[width=0.85\textwidth]{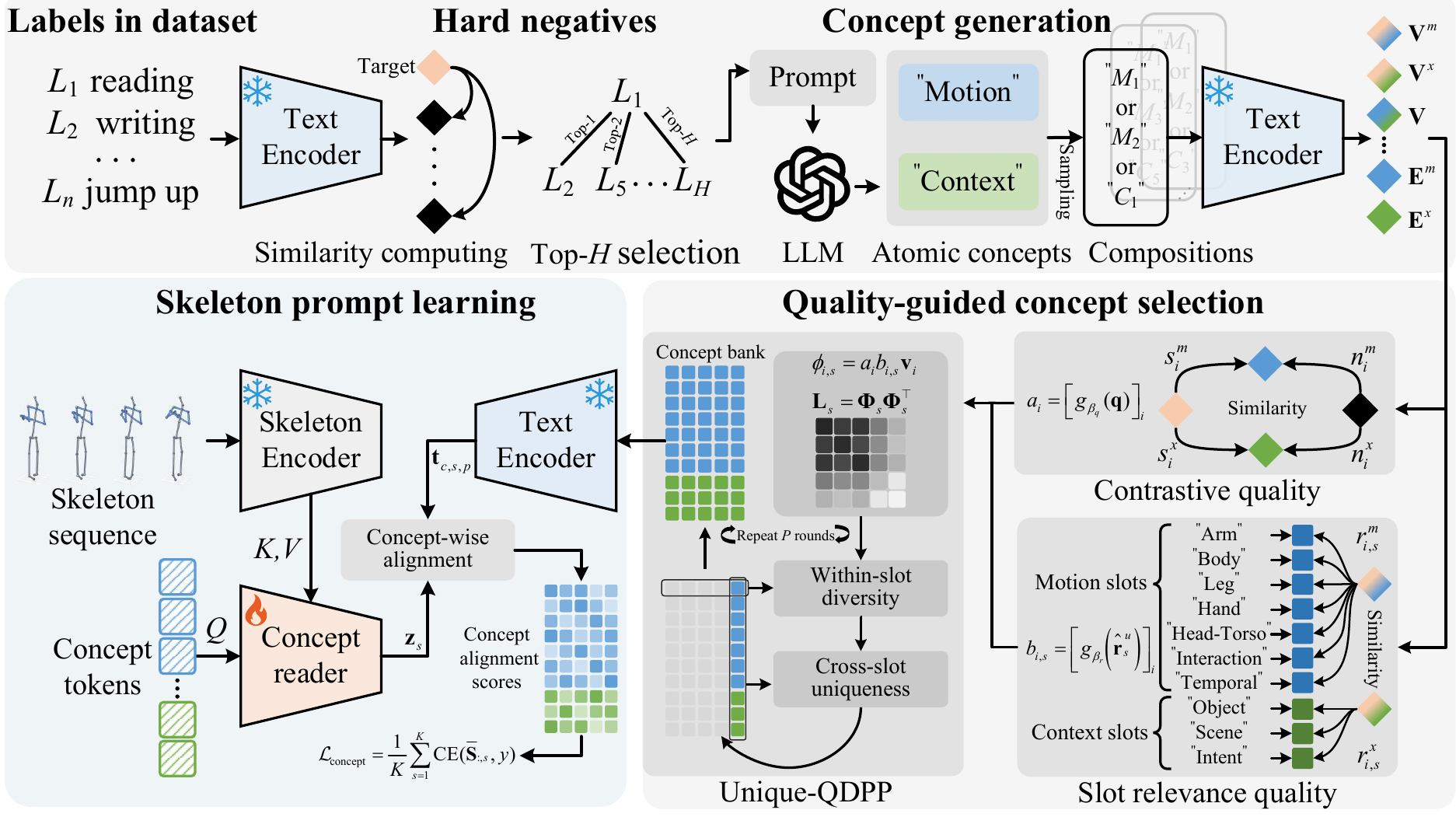}
\caption{Overview of \textbf{CRAFT}. Discriminative concept construction supplies slot-specific targets for skeleton prompt learning. These targets supervise concept tokens that read motion and context evidence from each input skeleton.}
\label{fig:main-framework}
\end{figure*}

\subsection{Skeleton-Text Alignment}

PURLS~\cite{zhu2024purls} learns adaptive skeleton groups, while STAR~\cite{chen2024star} uses visual-attribute and semantic-part prompts. SCoPLe~\cite{zhu2025scople} introduces joint-shaped skeleton prompts. STAR++~\cite{chen2026starpp} uses region-aware holistic fusion, and Neuron~\cite{chen2025neuron} learns evolving spatial-temporal micro-prototypes. SkeletonContext~\cite{wang2026skeletoncontext} links contextual semantics to skeleton features through masked contextual slot reconstruction. These methods structure skeleton-side learning through body parts, prompts, or reconstruction targets. \textbf{CRAFT} ties this structure to the selected concept bank. Supervision within each semantic slot trains input-dependent skeleton readouts to express its motion or context cues.


\section{Method}
\label{sec:method}

\paragraph{Framework overview.} \textbf{CRAFT} couples training-free discriminative concept construction with concept-supervised skeleton prompt learning (Fig.~\ref{fig:main-framework}). The first component selects complementary descriptions that distinguish confusable classes. The second uses this bank to supervise concept tokens that read evidence from each input skeleton.

\paragraph{Problem formulation.} Let $\mathcal C_{\mathrm{seen}}$ and $\mathcal C_{\mathrm{unseen}}$ denote disjoint sets of seen and unseen action classes, with $\mathcal C=\mathcal C_{\mathrm{seen}}\cup\mathcal C_{\mathrm{unseen}}$. Each sample $(\mathbf X,y)$ consists of a skeleton sequence $\mathbf X$ and its class label $y$. Action names $L_c$ are available for all classes $c\in\mathcal C$.
Under cross-subject evaluation, disjoint subject groups yield training and test pools $\mathcal D_{\mathrm{sub}}^{\mathrm{tr}}$ and $\mathcal D_{\mathrm{sub}}^{\mathrm{te}}$. We train on $\mathcal D_{\mathrm{tr}}^{\mathrm{seen}}=\{(\mathbf X,y)\in\mathcal D_{\mathrm{sub}}^{\mathrm{tr}}:y\in\mathcal C_{\mathrm{seen}}\}$. Seen- and unseen-class samples in $\mathcal D_{\mathrm{sub}}^{\mathrm{te}}$ form the test subsets $\mathcal D_{\mathrm{te}}^{\mathrm{seen}}$ and $\mathcal D_{\mathrm{te}}^{\mathrm{unseen}}$, respectively.

ZSL evaluates recognition on $\mathcal D_{\mathrm{te}}^{\mathrm{unseen}}$, whereas GZSL uses both test subsets, $\mathcal D_{\mathrm{te}}^{\mathrm{seen}}\cup\mathcal D_{\mathrm{te}}^{\mathrm{unseen}}$. The corresponding candidate class set is
\begin{equation}
\mathcal C_{\mathrm{test}}=\begin{cases}
\mathcal C_{\mathrm{unseen}},&\text{ZSL},\\
\mathcal C_{\mathrm{seen}}\cup\mathcal C_{\mathrm{unseen}},&\text{GZSL}.
\end{cases}
\label{eq:method-test-classes}
\end{equation}
Given a test sequence $\mathbf X$, the model assigns a score $f_c(\mathbf X)$ to each candidate class $c$ and predicts
\begin{equation}
\hat y=\operatorname*{arg\,max}_{c\in\mathcal C_{\mathrm{test}}}f_c(\mathbf X).
\label{eq:method-problem}
\end{equation}

\subsection{Discriminative Concept Construction}
\label{sec:concept-construction}

\paragraph{Hard-negative-conditioned generation.} Concept construction uses action names $L_c$ without skeleton samples. A frozen CLIP text encoder $E(\cdot)$~\cite{radford2021clip} returns $\ell_2$-normalized embeddings. Encoding each name with an action-name template gives $\mathbf t_c$. Cosine similarity retrieves the nearest non-target classes within the target's seen or unseen partition. We prompt the LLM with the target name and its $H$ nearest action names. This focuses the generated motion and context descriptions, or atoms, on differences among the closest rivals. String deduplication and cosine-similarity filtering at threshold $\delta_{\mathrm{dup}}$ yield pools of $M_m$ motion and $M_x$ context atoms. We sample $N$ distinct candidates from these pools. Each combines $R_m$ distinct motion atoms and $R_x$ distinct context atoms with the connective \texttt{or} inside an action-description template.

For candidate $i$, we encode its full description, motion component, and context component with the class-name composition template, yielding $\mathbf v_i$, $\mathbf v_i^m$, and $\mathbf v_i^x$. Generic prefixes for the same motion and context content yield \emph{evidence-only} embeddings $\mathbf e_i^m$ and $\mathbf e_i^x$. These prefixes omit $L_c$ from the template used for quality scoring. All five embeddings lie in $\mathbb R^d$, where $d$ is the text embedding dimension, and share candidate index $i\in\{1,\ldots,N\}$. We suppress class index $c$ for readability. Stacking the vectors as rows gives $\mathbf V$, $\mathbf V^m$, $\mathbf V^x$, $\mathbf E^m$, and $\mathbf E^x\in\mathbb R^{N\times d}$ in Fig.~\ref{fig:main-framework}. Component embeddings determine slot relevance; full-description embeddings measure diversity. The Appendix provides the LLM prompts and description templates.
\paragraph{Contrastive quality and slot relevance.} Contrastive quality measures how strongly a candidate's evidence favors its target over confusable classes. For motion or context component $u\in\{m,x\}$, target similarity is
\begin{equation}
s_i^u=(\mathbf e_i^u)^\top\mathbf t_c.
\label{eq:method-target-similarity}
\end{equation}
We pool similarities to the $H$ retrieved negative-class embeddings $\mathbf n_j$:
\begin{equation}
n_i^u=\tau_{\mathrm{hn}}\log\!\left[\frac1H\sum_{j=1}^{H}
\exp\!\left(\frac{(\mathbf e_i^u)^\top\mathbf n_j}{\tau_{\mathrm{hn}}}\right)\right].
\label{eq:method-negative-similarity}
\end{equation}
Here $\tau_{\mathrm{hn}}$ controls negative pooling. The combined contrastive quality is
\begin{equation}
q_i=\frac{(s_i^m-n_i^m)+\alpha(s_i^x-n_i^x)}{1+\alpha},
\label{eq:method-quality}
\end{equation}
where $\alpha$ balances motion and context. Larger $q_i$ indicates stronger separation from hard negatives. Generation and quality scoring use the same neighbor count $H$ (Sec.~\ref{sec:experimental-setup}).

Slot relevance measures how well a candidate matches a semantic role. We organize motion and context into $K$ shared slots. The anchor description $d_s^{\mathrm{anc}}$ specifies slot $s$, with embedding $\mathbf h_s=E(d_s^{\mathrm{anc}})$. The mapping $u(s)$ selects its motion or context component. Candidate relevance is
\begin{equation}
r_{i,s}=(\mathbf v_i^{u(s)})^\top\mathbf h_s.
\label{eq:method-relevance}
\end{equation}
Here $s\in\{1,\ldots,K\}$ indexes semantic slots. A hat denotes standardization over the current class's $N$ candidates, performed separately for each slot when computing $\hat r_{i,s}$. We floor the sample standard deviation at $\epsilon_{\mathrm{std}}$. The mapping $g_\beta(x)=\operatorname{clip}_{[w_{\min},w_{\max}]}(e^{\beta x})$ converts standardized scores into positive amplitudes. The parameter $\beta$ controls weighting strength, while $w_{\min}$ and $w_{\max}$ bound the amplitudes. Contrastive quality supplies the class-discriminative weight
\begin{equation}
a_i=g_{\beta_q}(\hat q_i),
\label{eq:method-quality-amplitude}
\end{equation}
while slot relevance supplies the slot-specific weight
\begin{equation}
b_{i,s}=g_{\beta_r}(\hat r_{i,s}).
\label{eq:method-relevance-amplitude}
\end{equation}

\paragraph{Unique-QDPP selection.} We develop Unique-QDPP by combining quality-weighted DPP (determinantal point process) selection~\cite{kulesza2012determinantal} within slots with uniqueness across slots. All slots draw from a shared pool of $N$ candidates per class. For slot $s$, contrastive quality and slot relevance jointly weight candidate $i$'s full-description embedding:
\begin{equation}
\boldsymbol{\phi}_{i,s}=a_i b_{i,s}\mathbf v_i.
\label{eq:method-weighted-feature}
\end{equation}
Stacking these vectors as rows of $\boldsymbol{\Phi}_s\in\mathbb R^{N\times d}$ gives the kernel
\begin{equation}
\mathbf L_s=\boldsymbol{\Phi}_s\boldsymbol{\Phi}_s^\top.
\label{eq:method-kernel}
\end{equation}
The kernel $\mathbf L_s\in\mathbb R^{N\times N}$ combines these weights with full-description similarity. Let $\mathcal S_s$ contain the indices already selected for slot $s$. Each remaining candidate receives the regularized DPP score
\begin{equation}
\begin{aligned}
\Delta_{i,s}={}&L_s(i,i)-\mathbf L_s[i,\mathcal S_s]\\[-2pt]
&\cdot\big(\mathbf L_s[\mathcal S_s,\mathcal S_s]+\epsilon_{\mathrm{dpp}}\mathbf I\big)^{-1}
\mathbf L_s[\mathcal S_s,i],
\end{aligned}
\label{eq:method-diversity}
\end{equation}
where $\mathbf I\in\mathbb R^{|\mathcal S_s|\times|\mathcal S_s|}$ is the identity matrix and $\epsilon_{\mathrm{dpp}}$ stabilizes the inverse. The second term discounts redundancy with descriptions already selected for this slot. It is zero when $\mathcal S_s$ is empty.

Independent selection within each slot can assign the same candidate to several slots. Such repetition reduces the number of distinct descriptions available under a fixed bank size. A class-wide exclusion set $\mathcal U$ enforces cross-slot uniqueness. It stores selected indices and indices of candidates with identical full text, ignoring case. All $\mathcal S_s$ and $\mathcal U$ start empty. Each round selects one candidate per slot: $i^*=\operatorname*{arg\,max}_{i\notin\mathcal U}\Delta_{i,s}$. We add $i^*$ to $\mathcal S_s$ and update $\mathcal U$ before visiting the next slot. The starting slot rotates across rounds to reduce selection-order bias.

After $P$ rounds, the bank $\mathcal B_c=\{d_{c,s,p}^{\mathrm{bank}}\}_{s,p}$ contains $P$ full descriptions per slot and $K\times P$ descriptions per class. The index $p\in\{1,\ldots,P\}$ enumerates selected descriptions within a slot. Slot type determines whether the motion or context component is used to assess relevance. Each selected description retains both components for reader supervision. Its embedding $\mathbf t_{c,s,p}=E(d_{c,s,p}^{\mathrm{bank}})$ supplies a slot-specific target for skeleton prompt learning. The weights $a_i b_{i,s}$ apply only during selection.

\subsection{Skeleton Prompt Learning}
\label{sec:skeleton-prompt-learning}

\paragraph{Concept-guided skeleton readout.} The concept bank supplies semantic targets for learnable skeleton prompts. We assign one token to each slot, forming $\mathbf Q\in\mathbb R^{K\times d}$. Row $s$ is initialized from the normalized mean of $\mathbf t_{c,s,p}$ over seen classes and selected descriptions. Tokens are shared across samples and classes, but their readouts depend on the input skeleton. Supervision within each slot trains its token to extract features associated with the slot's motion or context semantics. Seen- and unseen-class banks share the same slot definitions, allowing learned readouts to match unseen-class concepts at inference.

We obtain frozen skeleton features $\mathbf F_{\mathrm{sk}}\in\mathbb R^{N_{\mathrm{sk}}\times d_{\mathrm{sk}}}$ under the protocol in Sec.~\ref{sec:experimental-setup}, where $N_{\mathrm{sk}}$ is the number of tokens per reader input and $d_{\mathrm{sk}}$ is their feature dimension. The reader applies shared blocks to retained streams; a fused feature sequence forms a single input. With $\theta$ denoting all reader parameters, its output is
\begin{equation}
\mathbf Z=\operatorname{Reader}_\theta(\mathbf Q,\mathbf F_{\mathrm{sk}})
\in\mathbb R^{K\times d}.
\label{eq:method-reader}
\end{equation}
Each block uses multi-head attention (MHA)~\cite{vaswani2017attention}. The concept tokens $\mathbf Q$ initialize the query states. In head $h$, layer normalization (LN) and learned projections map these states to queries $\mathbf Q_h\in\mathbb R^{K\times d_h}$. Separate projections of normalized skeleton features produce keys and values $\mathbf K_h,\mathbf V_h\in\mathbb R^{N_{\mathrm{sk}}\times d_h}$, where $d_h$ is the head dimension. Each concept token attends to skeleton tokens with weights
\begin{equation}
\mathbf A_h=\operatorname{softmax}_{N_{\mathrm{sk}}}
\!\left(\frac{\mathbf Q_h\mathbf K_h^\top}{\sqrt{d_h}}\right).
\label{eq:method-attention-weights}
\end{equation}
Row $s$ of $\mathbf A_h\in\mathbb R^{K\times N_{\mathrm{sk}}}$ weights the skeleton tokens for concept token $s$. The weighted values give its readout:
\begin{equation}
\mathbf O_h=\mathbf A_h\mathbf V_h\in\mathbb R^{K\times d_h}.
\label{eq:method-attention-readout}
\end{equation}
We concatenate the head outputs and update the concept tokens through projection, residual connections, and a pre-LN feed-forward network (FFN). Subsequent blocks repeat this one-way readout over the same frozen skeleton tokens. If multiple streams are retained, learnable slot-specific softmax weights fuse their projected, $\ell_2$-normalized readouts. The normalized output $\mathbf Z\in\mathbb R^{K\times d}$ provides one input-dependent skeleton representation $\mathbf z_s$ per slot for alignment with the concept bank.

\paragraph{Concept-wise alignment and supervision.} We align each reader output $\mathbf z_s$ with full-description embeddings $\mathbf t_{c,s,p}$ from the same slot across classes:
\begin{equation}
S_{c,s,p}=\kappa\mathbf z_s^\top\mathbf t_{c,s,p},
\label{eq:method-matching}
\end{equation}
where both embeddings are unit-normalized and $\kappa$ is the exponential of the frozen CLIP logit-scale parameter. Averaging the $P$ description scores gives the slot-specific class logit:
\begin{equation}
\bar S_{c,s}=\frac1P\sum_{p=1}^{P}S_{c,s,p}.
\label{eq:method-proposal-average}
\end{equation}
We apply class supervision to each slot and average over the $K$ slots:
\begin{equation}
\mathcal L_{\mathrm{concept}}=
\mathbb E_{(\mathbf X,y)\sim\mathcal D_{\mathrm{tr}}^{\mathrm{seen}}}
\!\left[\frac1K\sum_{s=1}^{K}\operatorname{CE}(\bar{\mathbf S}_{:,s},y)\right].
\label{eq:method-concept-loss}
\end{equation}
Here $\bar{\mathbf S}_{:,s}$ contains seen-class logits for slot $s$, and CE denotes cross-entropy. This supervision trains the concept tokens and the reader to distinguish classes using the selected motion and context semantics.

\paragraph{Training and inference.} \textbf{CRAFT} replaces the baseline's action descriptions with the concept bank and adds a reader over frozen skeleton features. The resulting text and skeleton representations feed the baseline's recognition mechanism. We train the concept tokens and reader, including its projections and slot-specific fusion weights. The skeleton encoder, text encoder, and bank remain frozen. The objective is $\mathcal L=(1-\lambda)\mathcal L_{\mathrm{original}}+\lambda\mathcal L_{\mathrm{concept}}$, where $\mathcal L_{\mathrm{original}}$ is the baseline's loss and $\lambda\in[0,1]$ balances the objectives. At inference, shared slots match skeleton readouts to the bank of each candidate class $c\in\mathcal C_{\mathrm{test}}$. These matches contribute to the baseline's class score $f_c(\mathbf X)$ in Eq.~\eqref{eq:method-problem}. Each baseline-CRAFT pair uses the same GZSL calibration protocol. MSF~\cite{li2023msf}, Neuron~\cite{chen2025neuron}, and Flora~\cite{chen2026flora} retain their original mechanisms. Both TDSM~\cite{do2025tdsm} variants use Flora-style gating. Both PURLS~\cite{zhu2024purls} variants use Neuron/SkeletonContext-style calibration~\cite{chen2025neuron,wang2026skeletoncontext}.


\section{Experiments}
\label{sec:reported-benchmarks}
\label{sec:experiments}

\begin{table*}[!t]
\centering
\caption{Fixed-split ZSL top-1 accuracy (\%) under cross-subject evaluation. Baseline-CRAFT pairs use SynSE's Shift-GCN features~\cite{gupta2021synse}. Class splits follow~\cite{gupta2021synse}, with challenging splits from~\cite{zhu2024purls}. In Tables~\ref{tab:zsl-comparison}-\ref{tab:zsl-random-splits}, splits denote seen/unseen class counts, and parentheses show absolute accuracy gains over each host baseline. \textcolor{red}{\textbf{Red}}: best. \textcolor{blue}{\textbf{Blue}}: second-best. - denotes unreported results. SkContext denotes SkeletonContext.}
\label{tab:zsl-comparison}
\small
\setlength{\tabcolsep}{2.5pt}
\renewcommand{\arraystretch}{1.04}
\newsavebox{\craftZSLNatural}
\newcommand{\craftZSLRows}{%
\toprule
\multirow[c]{2}{*}[-2.5pt]{Method} & \multirow[c]{2}{*}[-2.5pt]{Pub.} & \multicolumn{4}{c}{NTU-60 (Acc., \%)} & \multicolumn{4}{c}{NTU-120 (Acc., \%)} \\
\cmidrule(lr){3-6}\cmidrule(lr){7-10}
 & & 55/5 & 48/12 & 40/20 & 30/30 & 110/10 & 96/24 & 80/40 & 60/60 \\
\midrule
ReViSE~\cite{tsai2017revise} & ICCV 17 & 53.9 & 17.5 & 24.3 & 14.8 & 55.0 & 32.4 & 19.5 & 8.3 \\
JPoSE~\cite{wray2019jpose} & ICCV 19 & 64.8 & 28.8 & 20.1 & 12.4 & 51.9 & 32.4 & 13.7 & 7.7 \\
CADA-VAE~\cite{schonfeld2019cadavae} & CVPR 19 & 76.8 & 29.0 & 16.2 & 11.5 & 59.5 & 35.8 & 10.6 & 5.7 \\
SynSE~\cite{gupta2021synse} & ICIP 21 & 75.8 & 33.3 & 19.9 & 12.0 & 62.7 & 38.7 & 13.6 & 7.7 \\
SMIE~\cite{zhou2023smie} & MM 23 & 78.0 & 40.2 & - & - & 65.7 & 45.3 & - & - \\
SA-DVAE~\cite{li2024sadvae} & ECCV 24 & 82.4 & 41.4 & - & - & 68.8 & 46.1 & - & - \\
STAR~\cite{chen2024star} & MM 24 & 81.4 & 45.1 & - & - & 63.3 & 44.3 & - & - \\
STAR++~\cite{chen2026starpp} & TCSVT 26 & 84.4 & 49.5 & - & - & 72.0 & 53.5 & - & - \\
DVTA~\cite{kuang2026dvta} & PR 26 & 79.3 & 44.1 & - & - & 74.9 & 51.8 & - & - \\
InfoCPL~\cite{xu2025infocpl} & TMM 25 & 85.9 & 53.3 & - & - & 74.8 & 60.1 & - & - \\
SCoPLe~\cite{zhu2025scople} & CVPR 25 & 84.1 & 53.0 & 32.0 & 18.2 & 74.5 & 52.2 & 25.3 & 15.7 \\
FS-VAE~\cite{wu2025fsvae} & ICCV 25 & 86.9 & 57.2 & - & - & 74.4 & 62.5 & - & - \\
SkContext~\cite{wang2026skeletoncontext} & CVPR 26 & \textcolor{red}{\textbf{89.6}} & 64.4 & 39.3 & 30.1 & 74.2 & 60.1 & 39.1 & 28.4 \\
\midrule
MSF~\cite{li2023msf} & ICIG 23 & 83.6 & 49.2 & 30.1 & 21.0 & 71.2 & 59.7 & 23.7 & 12.7 \\
\textbf{MSF+Ours} & This work & 83.9\,{\scriptsize (+0.3)} & 55.6\,{\scriptsize (+6.4)} & 37.3\,{\scriptsize (+7.2)} & 31.6\,{\scriptsize (+10.6)} & 75.1\,{\scriptsize (+3.9)} & 61.4\,{\scriptsize (+1.7)} & 37.2\,{\scriptsize (+13.5)} & 24.8\,{\scriptsize (+12.1)} \\
\midrule
PURLS~\cite{zhu2024purls} & CVPR 24 & 79.2 & 41.0 & 31.1 & 23.5 & 72.0 & 52.0 & 28.4 & 19.6 \\
\textbf{PURLS+Ours} & This work & 82.6\,{\scriptsize (+3.4)} & 56.1\,{\scriptsize (+15.1)} & 39.8\,{\scriptsize (+8.7)} & 33.8\,{\scriptsize (+10.3)} & 76.6\,{\scriptsize (+4.6)} & 63.5\,{\scriptsize (+11.5)} & 35.7\,{\scriptsize (+7.3)} & 26.4\,{\scriptsize (+6.8)} \\
\midrule
Neuron~\cite{chen2025neuron} & CVPR 25 & 86.9 & 62.7 & 36.6 & 27.2 & 71.5 & 57.1 & 21.8 & 16.2 \\
\textbf{Neuron+Ours} & This work & 87.4\,{\scriptsize (+0.5)} & \textcolor{blue}{\textbf{66.2}}\,{\scriptsize (+3.5)} & \textcolor{red}{\textbf{43.7}}\,{\scriptsize (+7.1)} & 32.6\,{\scriptsize (+5.4)} & 75.1\,{\scriptsize (+3.6)} & 61.8\,{\scriptsize (+4.7)} & 25.5\,{\scriptsize (+3.7)} & 23.6\,{\scriptsize (+7.4)} \\
\midrule
TDSM~\cite{do2025tdsm} & ICCV 25 & 86.5 & 56.0 & 36.1 & 25.9 & 74.2 & 65.1 & 37.0 & 27.2 \\
\textbf{TDSM+Ours} & This work & 87.1\,{\scriptsize (+0.6)} & 60.4\,{\scriptsize (+4.4)} & \textcolor{blue}{\textbf{41.8}}\,{\scriptsize (+5.7)} & \textcolor{blue}{\textbf{37.0}}\,{\scriptsize (+11.1)} & 77.0\,{\scriptsize (+2.8)} & \textcolor{red}{\textbf{69.2}}\,{\scriptsize (+4.1)} & \textcolor{red}{\textbf{44.3}}\,{\scriptsize (+7.3)} & 28.4\,{\scriptsize (+1.2)} \\
\midrule
Flora~\cite{chen2026flora} & CVPR 26 & 86.3 & 65.3 & 31.1 & 35.7 & \textcolor{blue}{\textbf{79.6}} & 66.4 & 40.1 & \textcolor{blue}{\textbf{29.0}} \\
\textbf{Flora+Ours} & This work & \textcolor{blue}{\textbf{87.5}}\,{\scriptsize (+1.2)} & \textcolor{red}{\textbf{66.6}}\,{\scriptsize (+1.3)} & 35.1\,{\scriptsize (+4.0)} & \textcolor{red}{\textbf{38.2}}\,{\scriptsize (+2.5)} & \textcolor{red}{\textbf{80.6}}\,{\scriptsize (+1.0)} & \textcolor{blue}{\textbf{68.3}}\,{\scriptsize (+1.9)} & \textcolor{blue}{\textbf{44.0}}\,{\scriptsize (+3.9)} & \textcolor{red}{\textbf{32.5}}\,{\scriptsize (+3.5)} \\
\bottomrule
}
\sbox{\craftZSLNatural}{\begin{tabular}{@{}lc*{8}{c}@{}}\craftZSLRows\end{tabular}}
\resizebox{\textwidth}{!}{%
\begin{tabular*}{\dimexpr\wd\craftZSLNatural*100/67\relax}{@{\extracolsep{\fill}}lc*{8}{c}@{}}
\craftZSLRows
\end{tabular*}%
}
\end{table*}

\begin{table}[!htbp]
\normalsize
\centering
\caption{Fixed-split GZSL results (\%) under cross-subject evaluation. Baseline-CRAFT pairs use SynSE's Shift-GCN features~\cite{gupta2021synse}. Class splits follow~\cite{li2023msf,chen2025neuron,chen2026flora,wang2026skeletoncontext}. $S$, $U$, and $H$ denote seen accuracy, unseen accuracy, and their harmonic mean.}
\label{tab:gzsl-comparison}
\fontsize{6.2}{7.5}\selectfont
\setlength{\tabcolsep}{1.3pt}
\renewcommand{\arraystretch}{1.08}
\begin{tabular*}{\columnwidth}{@{\extracolsep{\fill}}>{\fontsize{6}{7.2}\selectfont}l*{3}{c}@{\hspace{4pt}}*{3}{c}@{\hspace{5pt}}*{3}{c}@{\hspace{4pt}}*{3}{c}@{}}
\toprule
\multirow[c]{3}{*}[-3pt]{Method} & \multicolumn{6}{c}{NTU-60 (Acc., \%)} & \multicolumn{6}{c}{NTU-120 (Acc., \%)} \\
\cmidrule(lr){2-7}\cmidrule(lr){8-13}
 & \multicolumn{3}{c}{55/5} & \multicolumn{3}{c}{48/12} & \multicolumn{3}{c}{110/10} & \multicolumn{3}{c}{96/24} \\
\cmidrule(lr){2-4}\cmidrule(lr){5-7}\cmidrule(lr){8-10}\cmidrule(lr){11-13}
 & $S$ & $U$ & $H$ & $S$ & $U$ & $H$ & $S$ & $U$ & $H$ & $S$ & $U$ & $H$ \\
\midrule
ReViSE~\cite{tsai2017revise} & 40.8 & 50.2 & 45.0 & 21.8 & 14.8 & 17.6 & 0.6 & 14.5 & 1.1 & 3.4 & 1.5 & 2.1 \\
JPoSE~\cite{wray2019jpose} & 66.5 & 53.5 & 59.3 & 28.6 & 18.7 & 22.6 & 53.6 & 11.6 & 19.1 & 41.0 & 3.8 & 6.9 \\
CADA-VAE~\cite{schonfeld2019cadavae} & 56.1 & 56.0 & 56.0 & 50.4 & 25.0 & 33.4 & 50.2 & 43.9 & 46.8 & 48.3 & 27.5 & 35.1 \\
SynSE~\cite{gupta2021synse} & 51.3 & 47.4 & 49.2 & 44.1 & 22.9 & 30.1 & 57.3 & 43.2 & 49.5 & 48.1 & 32.9 & 39.1 \\
SA-DVAE~\cite{li2024sadvae} & 62.3 & 70.8 & 66.3 & 50.2 & 36.9 & 42.6 & 61.1 & 59.8 & 60.4 & 58.8 & 35.8 & 44.5 \\
STAR~\cite{chen2024star} & 69.0 & 69.9 & 69.4 & 62.7 & 37.0 & 46.6 & 59.9 & 52.7 & 56.1 & 51.2 & 36.9 & 42.9 \\
STAR++~\cite{chen2026starpp} & 61.1 & 73.6 & 66.8 & 58.2 & 40.4 & 47.7 & 59.0 & 55.4 & 57.2 & 52.8 & 45.2 & 48.7 \\
SCoPLe~\cite{zhu2025scople} & 69.6 & 71.9 & 70.8 & 54.5 & 61.8 & 57.9 & 63.5 & 61.1 & 62.3 & 53.3 & 51.2 & 52.2 \\
FS-VAE~\cite{wu2025fsvae} & 77.0 & 74.5 & 75.7 & 56.2 & 48.6 & 52.1 & 59.2 & 67.9 & 63.3 & 57.8 & 51.9 & 54.7 \\
SkContext~\cite{wang2026skeletoncontext} & 78.3 & 76.1 & 77.1 & 65.5 & 57.1 & 61.1 & 61.1 & 65.5 & 63.1 & 64.6 & 49.6 & \textcolor{blue}{\textbf{56.1}} \\
\midrule
MSF~\cite{li2023msf} & 71.7 & 66.2 & 68.8 & 58.8 & 40.0 & 47.6 & 46.8 & 68.3 & 55.6 & 56.8 & 48.6 & 52.4 \\
\textbf{MSF+Ours} & \begin{tabular}[c]{@{}c@{}}72.3\\{\fontsize{4.5}{5.2}\selectfont (+0.6)}\end{tabular} & \begin{tabular}[c]{@{}c@{}}71.8\\{\fontsize{4.5}{5.2}\selectfont (+5.6)}\end{tabular} & \begin{tabular}[c]{@{}c@{}}72.0\\{\fontsize{4.5}{5.2}\selectfont (+3.2)}\end{tabular} & \begin{tabular}[c]{@{}c@{}}60.4\\{\fontsize{4.5}{5.2}\selectfont (+1.6)}\end{tabular} & \begin{tabular}[c]{@{}c@{}}44.7\\{\fontsize{4.5}{5.2}\selectfont (+4.7)}\end{tabular} & \begin{tabular}[c]{@{}c@{}}51.3\\{\fontsize{4.5}{5.2}\selectfont (+3.7)}\end{tabular} & \begin{tabular}[c]{@{}c@{}}47.7\\{\fontsize{4.5}{5.2}\selectfont (+0.9)}\end{tabular} & \begin{tabular}[c]{@{}c@{}}69.6\\{\fontsize{4.5}{5.2}\selectfont (+1.3)}\end{tabular} & \begin{tabular}[c]{@{}c@{}}56.6\\{\fontsize{4.5}{5.2}\selectfont (+1.0)}\end{tabular} & \begin{tabular}[c]{@{}c@{}}61.4\\{\fontsize{4.5}{5.2}\selectfont (+4.6)}\end{tabular} & \begin{tabular}[c]{@{}c@{}}49.4\\{\fontsize{4.5}{5.2}\selectfont (+0.8)}\end{tabular} & \begin{tabular}[c]{@{}c@{}}54.7\\{\fontsize{4.5}{5.2}\selectfont (+2.3)}\end{tabular} \\
\midrule
PURLS~\cite{zhu2024purls} & 56.6 & 55.3 & 55.9 & 62.5 & 37.9 & 47.2 & 41.4 & 39.6 & 40.5 & 41.8 & 28.4 & 33.8 \\
\textbf{PURLS+Ours} & \begin{tabular}[c]{@{}c@{}}72.9\\{\fontsize{4.5}{5.2}\selectfont (+16.3)}\end{tabular} & \begin{tabular}[c]{@{}c@{}}57.2\\{\fontsize{4.5}{5.2}\selectfont (+1.9)}\end{tabular} & \begin{tabular}[c]{@{}c@{}}64.1\\{\fontsize{4.5}{5.2}\selectfont (+8.2)}\end{tabular} & \begin{tabular}[c]{@{}c@{}}65.5\\{\fontsize{4.5}{5.2}\selectfont (+3.0)}\end{tabular} & \begin{tabular}[c]{@{}c@{}}49.3\\{\fontsize{4.5}{5.2}\selectfont (+11.4)}\end{tabular} & \begin{tabular}[c]{@{}c@{}}56.2\\{\fontsize{4.5}{5.2}\selectfont (+9.0)}\end{tabular} & \begin{tabular}[c]{@{}c@{}}66.5\\{\fontsize{4.5}{5.2}\selectfont (+25.1)}\end{tabular} & \begin{tabular}[c]{@{}c@{}}66.5\\{\fontsize{4.5}{5.2}\selectfont (+26.9)}\end{tabular} & \begin{tabular}[c]{@{}c@{}}66.5\\{\fontsize{4.5}{5.2}\selectfont (+26.0)}\end{tabular} & \begin{tabular}[c]{@{}c@{}}59.5\\{\fontsize{4.5}{5.2}\selectfont (+17.7)}\end{tabular} & \begin{tabular}[c]{@{}c@{}}51.2\\{\fontsize{4.5}{5.2}\selectfont (+22.8)}\end{tabular} & \begin{tabular}[c]{@{}c@{}}55.0\\{\fontsize{4.5}{5.2}\selectfont (+21.2)}\end{tabular} \\
\midrule
Neuron~\cite{chen2025neuron} & 69.1 & 73.8 & 71.4 & 61.6 & 56.8 & 59.1 & 67.6 & 59.5 & 63.3 & 67.5 & 44.4 & 53.6 \\
\textbf{Neuron+Ours} & \begin{tabular}[c]{@{}c@{}}76.8\\{\fontsize{4.5}{5.2}\selectfont (+7.7)}\end{tabular} & \begin{tabular}[c]{@{}c@{}}75.4\\{\fontsize{4.5}{5.2}\selectfont (+1.6)}\end{tabular} & \begin{tabular}[c]{@{}c@{}}76.1\\{\fontsize{4.5}{5.2}\selectfont (+4.7)}\end{tabular} & \begin{tabular}[c]{@{}c@{}}65.1\\{\fontsize{4.5}{5.2}\selectfont (+3.5)}\end{tabular} & \begin{tabular}[c]{@{}c@{}}57.9\\{\fontsize{4.5}{5.2}\selectfont (+1.1)}\end{tabular} & \begin{tabular}[c]{@{}c@{}}\textcolor{blue}{\textbf{61.3}}\\{\fontsize{4.5}{5.2}\selectfont (+2.2)}\end{tabular} & \begin{tabular}[c]{@{}c@{}}68.7\\{\fontsize{4.5}{5.2}\selectfont (+1.1)}\end{tabular} & \begin{tabular}[c]{@{}c@{}}64.6\\{\fontsize{4.5}{5.2}\selectfont (+5.1)}\end{tabular} & \begin{tabular}[c]{@{}c@{}}66.6\\{\fontsize{4.5}{5.2}\selectfont (+3.3)}\end{tabular} & \begin{tabular}[c]{@{}c@{}}69.4\\{\fontsize{4.5}{5.2}\selectfont (+1.9)}\end{tabular} & \begin{tabular}[c]{@{}c@{}}46.4\\{\fontsize{4.5}{5.2}\selectfont (+2.0)}\end{tabular} & \begin{tabular}[c]{@{}c@{}}55.6\\{\fontsize{4.5}{5.2}\selectfont (+2.0)}\end{tabular} \\
\midrule
TDSM~\cite{do2025tdsm} & 71.8 & 69.5 & 70.6 & 60.1 & 37.1 & 45.8 & 57.6 & 59.4 & 58.4 & 54.7 & 50.2 & 52.3 \\
\textbf{TDSM+Ours} & \begin{tabular}[c]{@{}c@{}}72.8\\{\fontsize{4.5}{5.2}\selectfont (+1.0)}\end{tabular} & \begin{tabular}[c]{@{}c@{}}70.0\\{\fontsize{4.5}{5.2}\selectfont (+0.5)}\end{tabular} & \begin{tabular}[c]{@{}c@{}}71.3\\{\fontsize{4.5}{5.2}\selectfont (+0.7)}\end{tabular} & \begin{tabular}[c]{@{}c@{}}70.3\\{\fontsize{4.5}{5.2}\selectfont (+10.2)}\end{tabular} & \begin{tabular}[c]{@{}c@{}}42.9\\{\fontsize{4.5}{5.2}\selectfont (+5.8)}\end{tabular} & \begin{tabular}[c]{@{}c@{}}53.3\\{\fontsize{4.5}{5.2}\selectfont (+7.5)}\end{tabular} & \begin{tabular}[c]{@{}c@{}}70.9\\{\fontsize{4.5}{5.2}\selectfont (+13.3)}\end{tabular} & \begin{tabular}[c]{@{}c@{}}65.1\\{\fontsize{4.5}{5.2}\selectfont (+5.7)}\end{tabular} & \begin{tabular}[c]{@{}c@{}}\textcolor{red}{\textbf{67.9}}\\{\fontsize{4.5}{5.2}\selectfont (+9.5)}\end{tabular} & \begin{tabular}[c]{@{}c@{}}53.7\\{\fontsize{4.5}{5.2}\selectfont (-1.0)}\end{tabular} & \begin{tabular}[c]{@{}c@{}}58.7\\{\fontsize{4.5}{5.2}\selectfont (+8.5)}\end{tabular} & \begin{tabular}[c]{@{}c@{}}\textcolor{blue}{\textbf{56.1}}\\{\fontsize{4.5}{5.2}\selectfont (+3.8)}\end{tabular} \\
\midrule
Flora~\cite{chen2026flora} & 75.9 & 78.8 & \textcolor{blue}{\textbf{77.4}} & 63.7 & 57.5 & 60.5 & 66.2 & 66.0 & 66.1 & 55.9 & 50.7 & 53.2 \\
\textbf{Flora+Ours} & \begin{tabular}[c]{@{}c@{}}77.5\\{\fontsize{4.5}{5.2}\selectfont (+1.6)}\end{tabular} & \begin{tabular}[c]{@{}c@{}}81.6\\{\fontsize{4.5}{5.2}\selectfont (+2.8)}\end{tabular} & \begin{tabular}[c]{@{}c@{}}\textcolor{red}{\textbf{79.5}}\\{\fontsize{4.5}{5.2}\selectfont (+2.1)}\end{tabular} & \begin{tabular}[c]{@{}c@{}}68.4\\{\fontsize{4.5}{5.2}\selectfont (+4.7)}\end{tabular} & \begin{tabular}[c]{@{}c@{}}60.0\\{\fontsize{4.5}{5.2}\selectfont (+2.5)}\end{tabular} & \begin{tabular}[c]{@{}c@{}}\textcolor{red}{\textbf{63.9}}\\{\fontsize{4.5}{5.2}\selectfont (+3.4)}\end{tabular} & \begin{tabular}[c]{@{}c@{}}65.1\\{\fontsize{4.5}{5.2}\selectfont (-1.1)}\end{tabular} & \begin{tabular}[c]{@{}c@{}}69.5\\{\fontsize{4.5}{5.2}\selectfont (+3.5)}\end{tabular} & \begin{tabular}[c]{@{}c@{}}\textcolor{blue}{\textbf{67.3}}\\{\fontsize{4.5}{5.2}\selectfont (+1.2)}\end{tabular} & \begin{tabular}[c]{@{}c@{}}56.4\\{\fontsize{4.5}{5.2}\selectfont (+0.5)}\end{tabular} & \begin{tabular}[c]{@{}c@{}}55.9\\{\fontsize{4.5}{5.2}\selectfont (+5.2)}\end{tabular} & \begin{tabular}[c]{@{}c@{}}\textcolor{red}{\textbf{56.2}}\\{\fontsize{4.5}{5.2}\selectfont (+3.0)}\end{tabular} \\
\bottomrule
\end{tabular*}
\end{table}

\begin{table}[!t]
\normalsize
\centering
\caption{Random-split ZSL accuracy and GZSL $H$ (\%) under cross-subject evaluation. Features come from ST-GCN~\cite{yan2018spatial}; class splits follow~\cite{li2024sadvae,do2025tdsm,chen2026flora}. All results are averaged over the same three random class splits. TDSM GZSL baseline results are reproduced.}
\label{tab:zsl-random-splits}
\fontsize{7}{8}\selectfont
\setlength{\tabcolsep}{0.7pt}
\renewcommand{\arraystretch}{1.08}
\setlength{\aboverulesep}{1.2pt}
\setlength{\belowrulesep}{1.2pt}
\begin{tabular*}{\columnwidth}{@{\extracolsep{\fill}}l*{6}{c}@{}}
\toprule
\multirow[c]{2}{*}[-1pt]{Method} & \multicolumn{2}{c}{NTU-60 (55/5)} & \multicolumn{2}{c}{NTU-120 (110/10)} & \multicolumn{2}{c}{PKU-MMD (46/5)} \\
\cmidrule(lr){2-3}\cmidrule(lr){4-5}\cmidrule(lr){6-7}
 & ZSL & GZSL & ZSL & GZSL & ZSL & GZSL \\
\midrule
ReViSE~\cite{tsai2017revise} & 60.9 & 60.3 & 44.9 & 40.3 & 59.3 & 49.8 \\
JPoSE~\cite{wray2019jpose} & 59.4 & 60.1 & 46.7 & 43.7 & 57.2 & 51.6 \\
CADA-VAE~\cite{schonfeld2019cadavae} & 61.8 & 66.4 & 45.2 & 45.6 & 60.7 & 45.8 \\
SynSE~\cite{gupta2021synse} & 64.2 & 67.5 & 47.3 & 43.5 & 60.8 & 49.5 \\
SMIE~\cite{zhou2023smie} & 65.1 & - & 46.4 & - & 60.8 & - \\
SA-DVAE~\cite{li2024sadvae} & 84.2 & 75.3 & 50.7 & 47.5 & 66.5 & 54.7 \\
SCoPLe~\cite{zhu2025scople} & 83.7 & 77.7 & 53.3 & 54.1 & 71.4 & 54.9 \\
FS-VAE~\cite{wu2025fsvae} & - & - & - & - & 71.2 & 59.0 \\
\midrule
TDSM~\cite{do2025tdsm} & \textcolor{blue}{\textbf{88.9}} & 75.8 & 69.5 & 64.6 & 70.8 & 56.6 \\
\textbf{TDSM+Ours} & \textcolor{red}{\textbf{90.6}}\hspace{0.5pt}{\fontsize{5.5}{6}\selectfont (+1.7)} & 79.6\hspace{0.5pt}{\fontsize{5.5}{6}\selectfont (+3.8)} & 70.9\hspace{0.5pt}{\fontsize{5.5}{6}\selectfont (+1.4)} & \textcolor{blue}{\textbf{67.1}}\hspace{0.5pt}{\fontsize{5.5}{6}\selectfont (+2.5)} & \textcolor{red}{\textbf{73.2}}\hspace{0.5pt}{\fontsize{5.5}{6}\selectfont (+2.4)} & \textcolor{red}{\textbf{60.3}}\hspace{0.5pt}{\fontsize{5.5}{6}\selectfont (+3.7)} \\
\midrule
Flora~\cite{chen2026flora} & 88.6 & \textcolor{blue}{\textbf{80.2}} & \textcolor{blue}{\textbf{71.2}} & 63.0 & 71.6 & 59.5 \\
\textbf{Flora+Ours} & \textcolor{blue}{\textbf{88.9}}\hspace{0.5pt}{\fontsize{5.5}{6}\selectfont (+0.3)} & \textcolor{red}{\textbf{81.7}}\hspace{0.5pt}{\fontsize{5.5}{6}\selectfont (+1.5)} & \textcolor{red}{\textbf{72.2}}\hspace{0.5pt}{\fontsize{5.5}{6}\selectfont (+1.0)} & \textcolor{red}{\textbf{67.7}}\hspace{0.5pt}{\fontsize{5.5}{6}\selectfont (+4.7)} & \textcolor{blue}{\textbf{72.3}}\hspace{0.5pt}{\fontsize{5.5}{6}\selectfont (+0.7)} & \textcolor{blue}{\textbf{59.8}}\hspace{0.5pt}{\fontsize{5.5}{6}\selectfont (+0.3)} \\
\bottomrule
\end{tabular*}
\end{table}

\begin{figure*}[t]
\centering
    \includegraphics[width=0.85\textwidth]{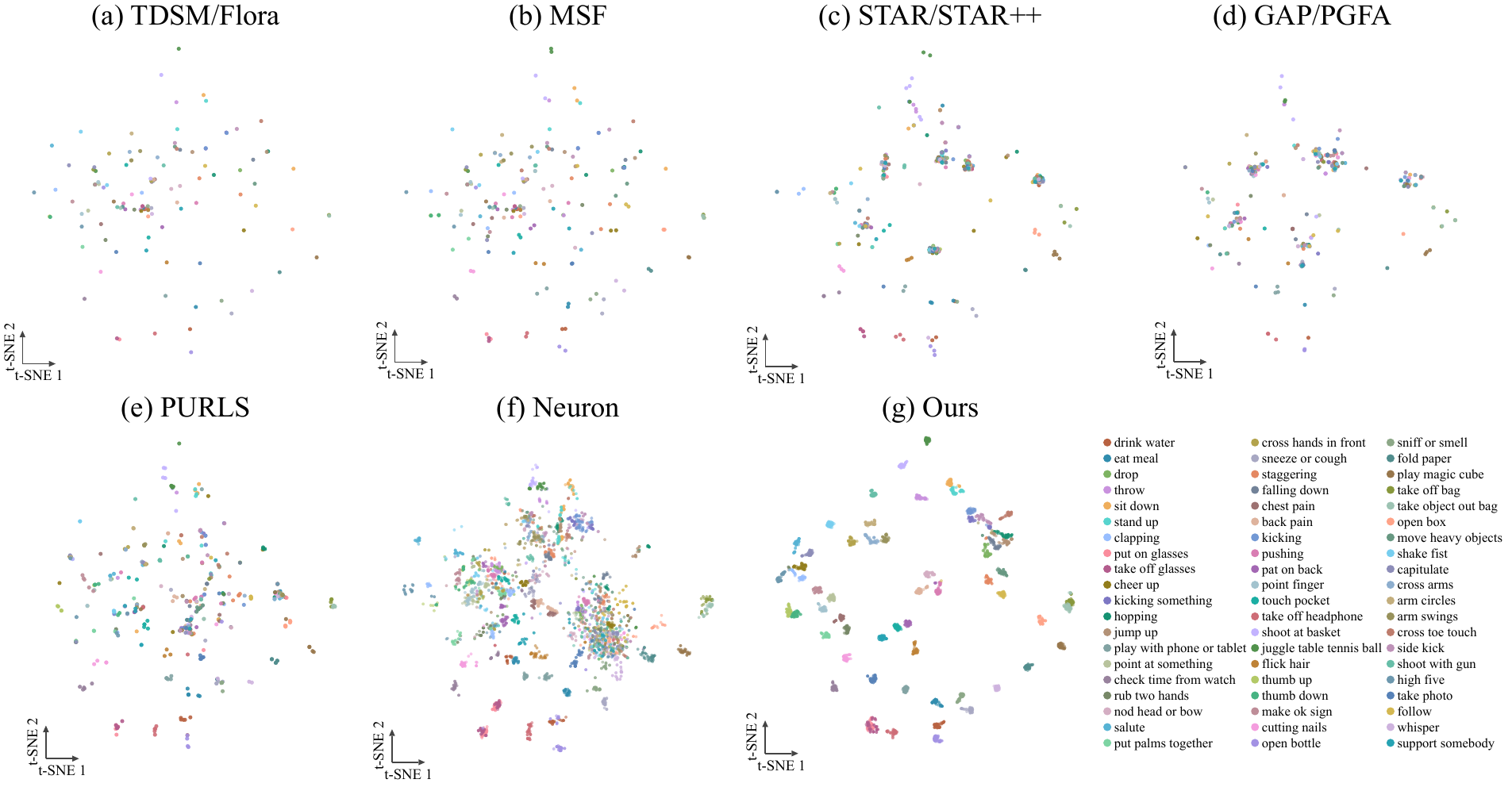}
    \caption{t-SNE visualization of action descriptions for the 60 unseen classes in the NTU-120 60/60 split. Colors denote the same classes across panels.}
\label{fig:text-bank-geometry}
\end{figure*}

\begin{table}[t]
\normalsize
\centering
\caption{Quantitative comparison of action descriptions. R@1/HNE: \%. MRR: $\times100$. \textcolor{red}{\textbf{Red}}: best. \textcolor{blue}{\textbf{Blue}}: second-best.}
\label{tab:text-asset-diagnostics}
\fontsize{7}{8.5}\selectfont
\setlength{\tabcolsep}{0.65pt}
\renewcommand{\arraystretch}{1.00}
\setlength{\aboverulesep}{1.2pt}
\setlength{\belowrulesep}{1.2pt}
\begin{tabular*}{\columnwidth}{@{\extracolsep{\fill}}>{\raggedright\arraybackslash}m{87pt}*{3}{r}@{\hspace{5pt}}*{3}{r}@{}}
\toprule
\multirow[c]{2}{*}[-1pt]{Method} & \multicolumn{3}{c}{NTU-60} & \multicolumn{3}{c}{NTU-120} \\
\cmidrule(lr){2-4}\cmidrule(lr){5-7}
 & {\fontsize{6.2}{7.4}\selectfont R@1$\uparrow$} & {\fontsize{6.2}{7.4}\selectfont MRR$\uparrow$} & {\fontsize{6.2}{7.4}\selectfont HNE$\downarrow$} & {\fontsize{6.2}{7.4}\selectfont R@1$\uparrow$} & {\fontsize{6.2}{7.4}\selectfont MRR$\uparrow$} & {\fontsize{6.2}{7.4}\selectfont HNE$\downarrow$} \\
\midrule
TDSM~\cite{do2025tdsm}/Flora~\cite{chen2026flora} & 36.7 & 52.1 & \textcolor{blue}{\textbf{26.7}} & \textcolor{blue}{\textbf{36.7}} & 51.2 & \textcolor{blue}{\textbf{35.8}} \\
MSF~\cite{li2023msf} & 23.3 & 40.7 & 45.0 & 22.5 & 36.5 & 50.0 \\
STAR~\cite{chen2024star}/STAR++~\cite{chen2026starpp} & 33.3 & 45.2 & 45.0 & 30.0 & 45.0 & 42.5 \\
GAP~\cite{xiang2023gap}/PGFA~\cite{zhou2025pgfa} & 6.7 & 19.6 & 65.0 & 7.5 & 16.9 & 70.0 \\
PURLS~\cite{zhu2024purls} & 36.7 & \textcolor{blue}{\textbf{53.3}} & 33.3 & \textcolor{blue}{\textbf{36.7}} & \textcolor{blue}{\textbf{51.4}} & 40.8 \\
Neuron~\cite{chen2025neuron} & \textcolor{blue}{\textbf{38.3}} & 53.0 & 36.7 & 30.0 & 43.6 & 48.3 \\
\midrule[0.25pt]
\textbf{Ours} & \textcolor{red}{\textbf{55.0}} & \textcolor{red}{\textbf{67.5}} & \textcolor{red}{\textbf{20.0}} & \textcolor{red}{\textbf{52.5}} & \textcolor{red}{\textbf{66.9}} & \textcolor{red}{\textbf{33.3}} \\
\bottomrule
\end{tabular*}
\end{table}

\subsection{Experimental Setup}
\label{sec:experimental-setup}

We evaluate fixed splits on NTU-60~\cite{shahroudy2016ntu} and NTU-120~\cite{liu2020ntu}, and random splits on both datasets and PKU-MMD~\cite{liu2017pkummd}. All evaluations use cross-subject protocols. Fixed splits follow prior ZSL~\cite{gupta2021synse,li2023msf,chen2025neuron,do2025tdsm,chen2026flora,wang2026skeletoncontext} and GZSL~\cite{li2023msf,chen2025neuron,chen2026flora,wang2026skeletoncontext} protocols, including challenging ZSL splits~\cite{zhu2024purls}. Random-split experiments follow~\cite{li2024sadvae,do2025tdsm,chen2026flora}, using the same three class splits, repetition counts, and averaging procedure for each baseline and its CRAFT variant. Each baseline-CRAFT pair uses the baseline's text encoder and shares skeleton features: SynSE~\cite{gupta2021synse} for fixed splits and ST-GCN~\cite{yan2018spatial} for random splits. We report ZSL top-1 accuracy and GZSL seen accuracy $S$, unseen accuracy $U$, and harmonic mean $H=2SU/(S+U)$.

We generate descriptions with \texttt{gpt-5.6-sol} and retain $M_m=35$ motion and $M_x=15$ context atoms per class, using deduplication threshold $\delta_{\mathrm{dup}}=0.95$. Each of the $N=500$ candidates combines $R_m=2$ motion atoms and $R_x=1$ context atom. We retain $P=5$ descriptions per slot across $K=10$ slots. Both ZSL and GZSL use separate seen and unseen banks, with hard negatives drawn within each partition. For fixed splits, generation and scoring share the same neighbor count $H$. Seen-class banks use $H=8$ on NTU-60 and $H=10$ on NTU-120. Smaller unseen partitions use fewer neighbors to focus generation and scoring on the closest confusable classes. In order of increasing unseen-class count, $H$ is $(2,3,5,6)$ on NTU-60 and $(3,5,8,10)$ on NTU-120. All five baselines and fixed splits share $\alpha=0.2$, $\tau_{\mathrm{hn}}=0.05$, $\beta_q=\beta_r=0.75$, $\epsilon_{\mathrm{std}}=10^{-6}$, $w_{\min}=0.25$, $w_{\max}=4$, $\epsilon_{\mathrm{dpp}}=10^{-5}$, and $\lambda=0.5$.

We reproduce each baseline's published ZSL scores with its released code before adding CRAFT. For PURLS and TDSM, we obtain unreported GZSL results using the calibration in Sec.~\ref{sec:skeleton-prompt-learning}. We train concept tokens and the reader on seen-class sequences from training subjects, keeping both encoders and banks frozen. Repeated training runs reuse the same preconstructed concept bank. MSF~\cite{li2023msf}, Neuron~\cite{chen2025neuron}, and PURLS~\cite{zhu2024purls} train for 50 epochs; TDSM~\cite{do2025tdsm} and Flora~\cite{chen2026flora} retain their original step budgets. On fixed and random splits, each baseline-CRAFT pair follows the baseline's optimization, model-selection, and evaluation settings, with GZSL calibration as above. PURLS provides the main ablations. The Appendix covers additional baseline ablations, parameter sensitivity, and multi-seed studies. All experiments were conducted on an NVIDIA RTX 5090 D v2 24GB GPU.

\subsection{Performance Comparison}
\label{sec:performance-comparison}

\paragraph{Fixed-split ZSL.}
\label{sec:zsl-results}
\textbf{CRAFT} improves all five baselines on all eight fixed NTU-60/120 splits (Table~\ref{tab:zsl-comparison}). Mean gains per baseline-dataset pair range from 2.3 to 9.4 percentage points (Fig.~\ref{fig:zsl-teaser}), showing broad applicability across baselines. CRAFT-enhanced models lead seven of the eight splits.
Gains are larger on challenging splits with fewer seen classes and more unseen classes. Averaged across baselines, they reach 7.3 percentage points on NTU-60 and 6.7 on NTU-120, versus 3.7 and 4.0 on the remaining splits. These gains show that class-discriminative concepts support generalization to a larger unseen vocabulary.

\paragraph{Fixed-split GZSL.}
\label{sec:gzsl-results}
GZSL is more demanding because unseen actions compete with seen classes. \textbf{CRAFT} improves unseen accuracy and $H$ on all 20 fixed baseline-split pairs (Table~\ref{tab:gzsl-comparison}). Mean $H$ gains range from 1.7 to 23.6 percentage points per baseline-dataset pair, and CRAFT-enhanced models lead all four splits. Joint gains in unseen accuracy and $H$ show effective recognition under this competition.

\paragraph{Random splits.}
\label{sec:random-split-results}
Random splits vary training and evaluation classes. \textbf{CRAFT} improves TDSM and Flora in every random-split ZSL and GZSL setting~\cite{li2024sadvae} across three datasets (Table~\ref{tab:zsl-random-splits}). Mean gains across baseline-dataset pairs are 1.3 percentage points in ZSL accuracy and 2.8 in GZSL $H$. On PKU-MMD, TDSM+Ours and Flora+Ours rank first and second in both metrics. These gains generalize across class partitions and datasets.

\paragraph{Action-description analysis.}
\label{sec:text-bank-analysis}
\label{sec:concept-bank-visualization}
We compare baseline descriptions and \textbf{CRAFT} concepts by text retrieval on NTU-60/120, using shared class-label queries and a frozen CLIP ViT-B/32 encoder~\cite{radford2021clip}. Table~\ref{tab:text-asset-diagnostics} reports recall at rank 1 (R@1)~\cite{faghri2018vsepp}, mean reciprocal rank (MRR)~\cite{voorhees1999trecqa}, and hard-neighbor error (HNE). For query $q_y$, class $c$ has score $s(q_y,c)=|T_c|^{-1}\sum_{t\in T_c}e(q_y)^\top e(t)$, where $T_c$ is its description set and $e(\cdot)$ gives unit-normalized embeddings. HNE measures confusion with $\mathcal N_3(y)$, the three nearest other class labels in CLIP space: $\mathrm{HNE}=\frac{100}{C}\sum_{y=1}^{C}\mathbf{1}\!\left[s(q_y,y)\leq\max_{c\in\mathcal N_3(y)}s(q_y,c)\right]$. Here $C$ counts classes and $\mathbf{1}[\cdot]$ denotes the indicator function. Against the strongest baseline for each metric, CRAFT improves R@1 by 16.7/15.8 percentage points and reduces HNE by 6.7/2.5 percentage points on NTU-60/120. We visualize the 60 unseen NTU-120 classes with openTSNE~\cite{policar2024opentsne} (Fig.~\ref{fig:text-bank-geometry}). All banks, including CRAFT, use identical centering, normalization, sample counts, and t-SNE settings. CRAFT concepts form compact, distinct class clusters with less overlap. These analyses support improved class discriminability in the text space. The Appendix outlines a skeleton-classification confusion analysis.

\subsection{Ablation study}
\label{sec:component-ablation}

\begingroup
\setlength{\intextsep}{5pt}
\begin{table}[!b]
\normalsize
\centering
\caption{Ablation study on PURLS (\%). Ablated models are retrained. Parentheses show changes from PURLS+Ours in percentage points.}
\label{tab:purls-ablation-accuracy}
\fontsize{7}{8}\selectfont
\setlength{\tabcolsep}{1pt}
\renewcommand{\arraystretch}{1.00}
\setlength{\aboverulesep}{1.2pt}
\setlength{\belowrulesep}{1.2pt}
\begin{tabular*}{\columnwidth}{@{\extracolsep{\fill}}lcccc@{}}
\toprule
\multicolumn{1}{@{}c}{\multirow{2}{*}{Variant}} & \multicolumn{2}{c}{NTU-60 (48/12)} & \multicolumn{2}{c}{NTU-120 (96/24)} \\
\cmidrule(lr){2-3}\cmidrule(lr){4-5}
 & ZSL & GZSL (\textit{H}) & ZSL & GZSL (\textit{H}) \\
\midrule

PURLS~\cite{zhu2024purls} & 41.0 & 47.2 & 52.0 & 33.8 \\
\textbf{PURLS+Ours} & \textbf{56.1} & \textbf{56.2} & \textbf{63.5} & \textbf{55.0} \\
\midrule[0.25pt]
\multicolumn{5}{@{}l}{\textit{A. Concept construction}} \\
w/o Concept bank & 45.9 \mbox{\fontsize{6}{8}\selectfont (-10.2)} & 48.4 \mbox{\fontsize{6}{8}\selectfont (-7.8)} & 45.2 \mbox{\fontsize{6}{8}\selectfont (-18.3)} & 46.9 \mbox{\fontsize{6}{8}\selectfont (-8.1)} \\
w/o Hard negatives & 53.1 \mbox{\fontsize{6}{8}\selectfont (-3.0)} & 52.6 \mbox{\fontsize{6}{8}\selectfont (-3.6)} & 59.4 \mbox{\fontsize{6}{8}\selectfont (-4.1)} & 50.8 \mbox{\fontsize{6}{8}\selectfont (-4.2)} \\
w/o Unique-QDPP & 45.2 \mbox{\fontsize{6}{8}\selectfont (-10.9)} & 45.7 \mbox{\fontsize{6}{8}\selectfont (-10.5)} & 57.7 \mbox{\fontsize{6}{8}\selectfont (-5.8)} & 53.6 \mbox{\fontsize{6}{8}\selectfont (-1.4)} \\
w/o Within-slot DPP Diversity & 53.7 \mbox{\fontsize{6}{8}\selectfont (-2.4)} & 52.1 \mbox{\fontsize{6}{8}\selectfont (-4.1)} & 59.3 \mbox{\fontsize{6}{8}\selectfont (-4.2)} & 54.1 \mbox{\fontsize{6}{8}\selectfont (-0.9)} \\
w/o Cross-slot Uniqueness & 51.0 \mbox{\fontsize{6}{8}\selectfont (-5.1)} & 49.4 \mbox{\fontsize{6}{8}\selectfont (-6.8)} & 58.4 \mbox{\fontsize{6}{8}\selectfont (-5.1)} & 53.9 \mbox{\fontsize{6}{8}\selectfont (-1.1)} \\
w/o Contrastive quality & 50.5 \mbox{\fontsize{6}{8}\selectfont (-5.6)} & 51.9 \mbox{\fontsize{6}{8}\selectfont (-4.3)} & 60.1 \mbox{\fontsize{6}{8}\selectfont (-3.4)} & 51.0 \mbox{\fontsize{6}{8}\selectfont (-4.0)} \\
w/o Slot relevance & 52.2 \mbox{\fontsize{6}{8}\selectfont (-3.9)} & 51.7 \mbox{\fontsize{6}{8}\selectfont (-4.5)} & 59.3 \mbox{\fontsize{6}{8}\selectfont (-4.2)} & 53.7 \mbox{\fontsize{6}{8}\selectfont (-1.3)} \\
\midrule[0.25pt]
\multicolumn{5}{@{}l}{\textit{B. Skeleton prompt learning}} \\
w/o Concept reader & 47.9 \mbox{\fontsize{6}{8}\selectfont (-8.2)} & 49.2 \mbox{\fontsize{6}{8}\selectfont (-7.0)} & 50.4 \mbox{\fontsize{6}{8}\selectfont (-13.1)} & 48.6 \mbox{\fontsize{6}{8}\selectfont (-6.4)} \\
w/o Concept loss & 51.8 \mbox{\fontsize{6}{8}\selectfont (-4.3)} & 55.3 \mbox{\fontsize{6}{8}\selectfont (-0.9)} & 54.0 \mbox{\fontsize{6}{8}\selectfont (-9.5)} & 41.3 \mbox{\fontsize{6}{8}\selectfont (-13.7)} \\
w/o Motion slots & 38.0 \mbox{\fontsize{6}{8}\selectfont (-18.1)} & 32.6 \mbox{\fontsize{6}{8}\selectfont (-23.6)} & 42.6 \mbox{\fontsize{6}{8}\selectfont (-20.9)} & 18.3 \mbox{\fontsize{6}{8}\selectfont (-36.7)} \\
w/o Context slots & 51.2 \mbox{\fontsize{6}{8}\selectfont (-4.9)} & 50.8 \mbox{\fontsize{6}{8}\selectfont (-5.4)} & 51.6 \mbox{\fontsize{6}{8}\selectfont (-11.9)} & 31.5 \mbox{\fontsize{6}{8}\selectfont (-23.5)} \\
\midrule[0.25pt]
\multicolumn{5}{@{}l}{\textit{C. Selection and slot-assignment controls}} \\
Random description selection & 43.4 \mbox{\fontsize{6}{8}\selectfont (-12.7)} & 47.6 \mbox{\fontsize{6}{8}\selectfont (-8.6)} & 53.6 \mbox{\fontsize{6}{8}\selectfont (-9.9)} & 45.8 \mbox{\fontsize{6}{8}\selectfont (-9.2)} \\
Within-class slot shuffle & 47.3 \mbox{\fontsize{6}{8}\selectfont (-8.8)} & 48.7 \mbox{\fontsize{6}{8}\selectfont (-7.5)} & 55.8 \mbox{\fontsize{6}{8}\selectfont (-7.7)} & 49.0 \mbox{\fontsize{6}{8}\selectfont (-6.0)} \\
\bottomrule
\end{tabular*}
\end{table}
\endgroup

\paragraph{Discriminative concept construction.} Every construction ablation in Table~\ref{tab:purls-ablation-accuracy} lowers ZSL accuracy and GZSL $H$ on both 48/12 and 96/24. Replacing the bank with PURLS's original descriptions (\emph{w/o Concept bank}) causes the largest mean ZSL drop among these variants, 14.3 percentage points. Mean $H$ drops by 8.0 percentage points. The retained tokens recover only part of the full model's gains with these original descriptions. Random selection costs 11.3/8.9 percentage points in mean ZSL accuracy/GZSL $H$; slot-relevance selection (\emph{w/o Unique-QDPP}) costs 8.4/6.0 percentage points. Removing hard-negative conditioning costs 3.6/3.9 percentage points, and removing contrastive quality also reduces both metrics. Class comparison thus contributes during both generation and scoring. Disabling either within-slot DPP diversity or cross-slot uniqueness lowers both metrics on both splits. Removing cross-slot uniqueness has the larger effect in all four comparisons.

\paragraph{Concept-supervised skeleton prompt learning.} Unlike random selection, within-class slot shuffling preserves descriptions but changes their assignment. Shuffling lowers mean ZSL accuracy/GZSL $H$ by 8.3/6.8 percentage points, supporting consistent slot assignment alongside selected content. Removing the reader or concept loss costs 6.9-10.7/6.7-7.3 percentage points in mean ZSL accuracy/GZSL $H$. Without the concept bank, reader outputs align with original text features. Without the reader, baseline skeleton features align with concept features. On 96/24, these variants score 45.2\% and 50.4\%, below PURLS's 52.0\% ZSL accuracy. Full CRAFT reaches 63.5\%. Removing seven motion slots costs 19.5/30.2 percentage points in mean ZSL accuracy/GZSL $H$; removing three context slots costs 8.4/14.5. Both groups contribute, with larger losses from removing motion slots.

\begin{figure}[!t]
\centering
\includegraphics[width=0.85\columnwidth]{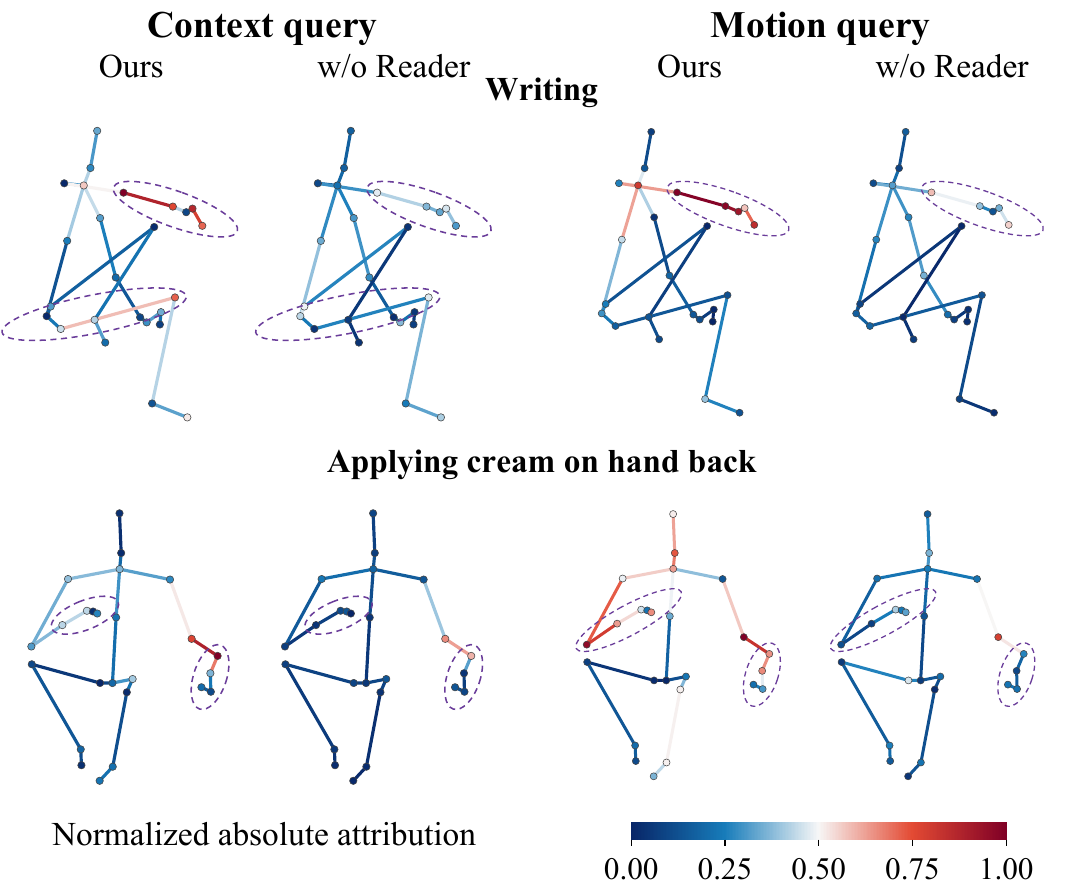}
\caption{Absolute integrated-gradient attribution for context and motion queries. Each model pair shares a maximum-normalized scale (blue: low; red: high).}
\label{fig:context-grounding}
\end{figure}

\begin{table}[!t]
\centering
\caption{Motion and context content from retained descriptions for Fig.~\ref{fig:context-grounding}. Queries use the motion and first context entries.}
\label{tab:context-descriptions}
\fontsize{7}{8}\selectfont
\setlength{\tabcolsep}{2pt}
\setlength{\aboverulesep}{1.2pt}
\setlength{\belowrulesep}{1.2pt}
\renewcommand{\arraystretch}{1.04}
\begin{tabular}{@{}>{\raggedright\arraybackslash}p{0.13\columnwidth}>{\raggedright\arraybackslash}p{\dimexpr0.87\columnwidth-4pt\relax}@{}}
\toprule
Type & \actionlabel{writing} (unseen)\\
\midrule
Motion & \actiondescription{Writing-side elbow stays bent near the torso while the forearm subtly pivots.}\\
Context & \actiondescription{A slender pen or pencil is held in the active hand.}\\
 & \actiondescription{The action typically occurs at a desk or table, often from a seated position.}\\
 & \actiondescription{A desk, table, or lap-based seated workspace may support the writing activity.}\\
\midrule
Type & \actionlabel{applying cream on hand back} (seen)\\
\midrule
Motion & \actiondescription{The applying hand performs small circular or back-and-forth strokes localized over the hand dorsum.}\\
Context & \actiondescription{A cream or lotion container may be visible or handled immediately before the rubbing phase.}\\
 & \actiondescription{The scene suggests solitary skin-care, with no recipient or exchanged object.}\\
 & \actiondescription{The apparent intent is skincare application rather than signaling, clapping, or object transfer.}\\
\bottomrule
\end{tabular}
\end{table}

\subsection{Concept Grounding Analysis}
\label{sec:context-grounding}

On NTU-120 96/24, Fig.~\ref{fig:context-grounding} compares PURLS+Ours with retrained \emph{w/o Concept reader} on the same skeleton inputs and queries (Table~\ref{tab:context-descriptions}). The full model's query score combines contributions from baseline skeleton features and reader outputs across all concept slots, including the three context slots. Features come from the final Shift-GCN block before global pooling. We average over people, fuse the four streams, and pool time into three bins. The resulting 75 skeleton tokens preserve the indices of the 25 joints. We compute integrated gradients~\cite{sundararajan2017axiomatic} relative to a spatial-mean reference, then sum signed attributions over time and channels at each joint index. Magnitudes are mapped to the original joints and share a color scale within each model pair. Each displayed joint value summarizes attribution to a deep feature that integrates information across joints and time. Dashed ellipses mark corresponding body regions.

In these examples, the reader shifts motion-query attribution toward the writing-side elbow and forearm, and the arms and hands applying cream. The hand-node share of total absolute attribution rises from 38.0\% to 44.0\% for writing and from 33.3\% to 35.6\% for cream application. Under context queries, it rises from 31.7\% to 35.9\% and from 37.0\% to 42.9\%, respectively. Writing's lower-body context attribution is consistent with its seated posture and the retained desk or seated-workspace descriptions (Table~\ref{tab:context-descriptions}). These examples illustrate how the reader connects selected concepts to local movement and posture, complementing the recognition evidence from retrained ablations.


\FloatBarrier
\section{Conclusion}
\label{sec:conclusion}

\textbf{CRAFT} builds discriminative concepts to supervise skeleton prompt learning, with gains on five baselines and three datasets. These results demonstrate CRAFT's effectiveness and compatibility. A broader challenge for skeleton encoders is their limited pretraining scale, which may constrain generalization relative to CLIP's encoders. More generalizable encoders could yield further gains with CRAFT's concept-guided learning.

{
    \small
    \bibliographystyle{ieeenat_fullname}
    \bibliography{references}
}
\end{document}